\documentclass{article}

\usepackage{arxiv}

\usepackage[utf8]{inputenc} % allow utf-8 input
\usepackage[T1]{fontenc}    % use 8-bit T1 fonts
\usepackage{hyperref}       % hyperlinks
\usepackage{url}            % simple URL typesetting
\usepackage{booktabs}       % professional-quality tables
\usepackage{amsfonts}       % blackboard math symbols
\usepackage{nicefrac}       % compact symbols for 1/2, etc.
\usepackage{microtype}      % microtypography

\usepackage[ruled,lined]{algorithm2e}% provides Algorithm environment
\usepackage{graphicx}% allows for inclusion of graphic files (figures)
\usepackage{blindtext}
\usepackage[T1]{fontenc}
\usepackage[utf8]{inputenc}
\usepackage{float}
\usepackage{todonotes}
\usepackage{enumitem}

\setcitestyle{square,numbers}
\title{Comixify: Transform video into a comics}

\author{
  Maciej Pęśko, Adam Svystun, Paweł Andruszkiewicz, Przemysław Rokita and Tomasz Trzciński \\
  Faculty of Electronics and Information Technology \\
  Warsaw University of Technology \\ 
  Nowowiejska 15/19, 00-665 Warszawa, Poland \\
  \texttt{243678{@}pw.edu.pl, 280814{@}pw.edu.pl, 261406{@}pw.edu.pl,  pro{@}ii.pw.edu.pl, t.trzcinski{@}ii.pw.edu.pl} \\
}

\begin{document}
\maketitle

\begin{abstract}
  In this paper, we propose a solution to transform a video into a comics. We approach this task using a neural style algorithm based on Generative Adversarial Networks (GANs). Several recent works in the field of Neural Style Transfer showed that producing an image in the style of another image is feasible. In this paper, we build up on these works and extend the existing set of style transfer use cases with a working application of video comixification. To that end, we train an end-to-end solution that transforms input video into a comics in two stages. In the first stage, we propose a state-of-the-art keyframes extraction algorithm that selects a subset of frames from the video to provide the most comprehensive video context and we filter those frames using image aesthetic estimation engine. In the second stage, the style of selected keyframes is transferred into a comics. To provide the most aesthetically compelling results, we selected the most state-of-the art style transfer solution and based on that implement our own ComixGAN framework. The final contribution of our work is a Web-based working application of video comixification available at \url{http://comixify.ii.pw.edu.pl}.
\end{abstract}

\keywords{Neural Style Transfer, Style Transfer, Comics Style, Comics, Computer Vision, Neural Network}

\section{Introduction}

\begin{figure}[t!]
  \centering
  \includegraphics[width=1.0\textwidth]{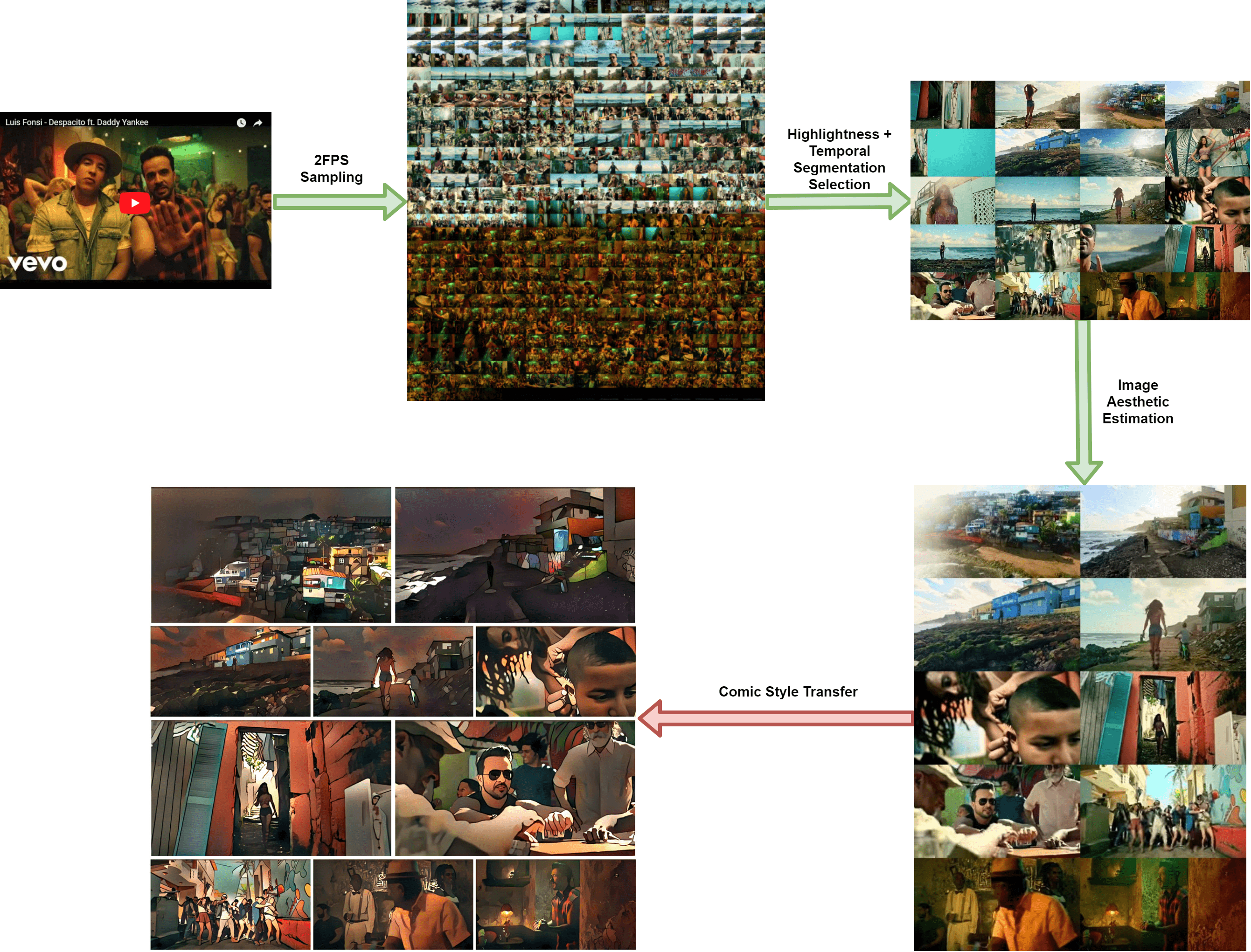}
  \caption{Comixify pipeline schema. A video is first processed by our keyframe extractor (green arrows). After sampling video with 2fps frequency, we select subset of frames that represents the video well, using highlightness score and temporal segmentation of the video. Afterwards selected frames are evaluated using image aesthetic estimation engine - to narrow down the number of extracted keyframes. Second stage (red arrow) is comic style transfer.\label{pipeline-schema} }
\end{figure}

Cartoons and comics became a popular mean of artistic expression worldwide. Unfortunately, only a limited number of talented people with painting or graphics skills can create them holding on to aesthetic standards. What is more, it also takes a significant amount of time to create a valuable comics graphics. Providing an automatic tool that converts videos to comics could revolutionize the way publishers and individuals create {\it comixified} content.

To advance the development of such tool we propose a working solution for the task of video comixification. We split this problem into two separate ones: (a) frame extraction and (b) style transfer. The goal of first stage is to select a subset of frames from video which provides the most comprehensive video context, being also visually attractive to viewer of the comics. In this stage we propose to use a state-of-the-art keyframe extraction algorithm based on reinforcement learning, which we further extend by combining temporal segmentation method with the  image aesthetic estimation. In the second stage, we transform the style of selected keyframes into a comics. To achieve the best results we analyze the existing style transfer approaches on the specific problem of comics style transfer. We compare their advantages and disadvantages in terms of qualitative and quantitative analysis, to find the most aesthetically pleasing solution. The final contribution of our work is a web-based working application of video comixification available at  \url{http://comixify.ii.pw.edu.pl}.

The remainder of this paper is organized in the following manner. In Section 2 we describe all related works that discuss the topic of style transfer and key frames extraction. Section 3 describes the details of the method proposed for video comixification.
Section 4 shows the details about our web-based application. In Section 5 we make some conclusions and plans for further research.

\section{Related Work}

\subsection{Keyframes Extraction}

The task of keyframes extraction is generally similar to the video summarization task. Both of them try to find a subset of input frames that provides a comprehensive representatation of the video. In the recent years  video summarization has gained a significant amount of attention from the research community. This attention may have been sparked by the availability of benchmark datasets, such as SumMe \cite{gygli2014} and TVSum \cite{song2015}. Zhang \textit{et al.} \cite{zhang2016} used Long Short-Term Memory (LSTM) to model the variable-range temporal dependency among video frames, so as to derive both representative and compact video summaries. Later, Mahasseni \textit{et al.} \cite{mahasseni2017} achieved even better results by using novel generative adversarial framework (GAN), consisting of summarizer and discriminator. The proposed summarizer was implemented as an autoencoder LSTM network with the objective of, first, selecting video frames, and then decoding the obtained summarization for reconstructing the input video. Discriminator was another LSTM aimed at distinguishing between the original video and its reconstruction from the summarizer. Lastly, Zhou \textit{et al.} \cite{zhou2017reinforcevsumm} proposed an end-to-end, reinforcement learning based framework, where they designed a novel reward function that jointly accounts for diversity and representativeness of generated summaries. Extensive experiments on two benchmark datasets showed that this unsupervised method not only outperforms other state-of-the-art unsupervised methods, but also is comparable or even superior to most of supervised approaches. Contrary to the above proposed methods, in this paper we leverage keyframe extraction and extend it with aesthetic estimation.  For this task, we propose two approaches - our popularity estimation engine and image quality estimation model.

Similarly to video summarization, a considerable amount of research was carried out in the domain of content popularity estimation. Various aspects of popularity prediction were looked at such as analyzing the user-access patterns relation to popularity \cite{almeida1996, chesire2001},  prediction of  the peak popularity time \cite{jiang2014} or popularity evolution patterns \cite{crane2008, szabo2010, pinto2013}. Khosla \textit{et al.} \cite{khosla2014} proposed using visual cues in the context of popularity prediction. Authors train Support Vector Machines models using among others deep neural network outputs. Trzciński {\it et al.} used a similar approach as the baseline model in \cite{popularity}, where authors proposed the use of recurrent neural networks for video popularity prediction.

Image quality assessment is also a popular topic among researchers. Large datasets for aesthetic visual analysis are available, such as AVA \cite{murray2012} and TID2013 \cite{ponomarenko2013}. Many studies \cite{lu2015, kao2015, kim2017} have proved that the extraction of high-level features using convolutional neural networks (CNNs) can result in significantly better results in image quality assessment task than previous methods, based on hand-crafted features, like in \cite{murray2012}. Talebi and Milanfar \cite{talebi2018} proposed a novel approach, based on a prediction of the distribution of human opinion scores. They used the squared EMD (earth mover's distance) loss proposed in \cite{hou2016}. They achieved results close to the state of the art, using much less computationally expensive model. In this work we combine the advancements in video summarization and popularity estimation, to produce a keyframe extraction solution most suitable for the keyframe extraction task.

\subsection{Style Transfer}

Gatys \textit{et al.} \cite{gatys:} in their paper demonstrated that deep convolutional neural networks are able to encode the style information of any given image. Moreover, they showed that the content and style of the image can be separated and treated individually. As a consequence, it is possible to transfer the characteristics of the style of a given image to another one, while preserving the content of the latter. They proposed to exploit the correlations between the features of deep neural networks in terms of Gram matrices to capture image style. Furthermore, Y. Li \textit{et al.} showed that covariance matrix can be as effective as Gram matrix in representing image style \cite{cov:gram} and theoretically proved that matching the Gram matrices of the neural activations can be seen as minimizing a specific Maximum Mean Discrepancy \cite{demystif:} function, which gives more intuition on why Gram matrix can represent an artistic style.

Since the work of \cite{gatys:}, numerous  improvements have been made in the field of Style Transfer. Johnson \textit{et al.} \cite{johnson:} and Ulyanov \textit{et al.} \cite{ulyanov:1} proposed fast approaches that increase the efficiency of style transfer by three orders of magnitude. Nevertheless, this improvement comes at a price of lower results quality. Multiple authors tried to address this shortcoming~ \cite{ulyanov:2,inter:layer,multimodal:,histogram:losses} or make it more generic to enable using different styles in one model. The proposed solution include using a conditional normalization layer that learns normalization parameters for each style \cite{dumoulin:}, swapping a given content feature with the closest style feature in its local neighborhood~\cite{chen:schmidt}, directly adjusting the content feature to match the mean and variance of the style feature \cite{huang:belongie,desai:,ghiasi:} or building a meta network which takes in the style image and produces corresponding image transformation network directly \cite{meta:net}. Other methods include an automatic segmentation of the objects and extraction of their soft semantic masks \cite{zao:rosik} or adjusting feature maps using whitening and coloring transforms \cite{universal:style}. There are also some works that try to make photo-realistic style transfer \cite{photo:real,deep:photo} to carry style of one photo to another, leaving it as realistic as possible. In addition, many works have been created that focus on some other, various fields of Style Transfer. For instance Coherent Online Video Style Transfer \cite{chen2017coherent} which describes end-to-end network that generates consistent stylized video sequences in near real time, StyleBank \cite{chen2017stylebank} which uses multiple convolution filter banks, where each filter in bank explicitly represents one style or Stereoscopic Neural Style Transfer \cite{chen2018stereoscopic} that concerns on 3D or AR/VR subject. Recently, Chen, Yang et al. in \cite{cartoongan} presented style transfer method based on Generative Adversarial Networks that seems to work really well in terms of photo cartoonization problem.

It is worth mentioning that existing approaches often suffer from a trade-off between generalization, quality and efficiency. More precisely, the optimization-based approaches handle arbitrary styles with a great visual quality, but the computational costs are relatively high. On the other side, the feed-forward methods are executed very fast with slightly worse but acceptable quality, but they are limited to a specific, fixed number of styles. Finally, the arbitrary methods are fast and enable multiple style transfer, but often their quality is worse than the previously proposed ones.

\subsection{Video comixification}

The topic of video comixification has also gained some interest in the industry - Google Research has recently launched their Android application called Storyboard~\footnote{\url{https://play.google.com/store/apps/details?id=com.google.android.apps.photolab.storyboard}} which allows you to transform videos into comics on your mobile device. Contrary to the solution provided by Google, we allow the user to select a few methods for keyframes extraction and Style Transfer and enable selection of not only the most comprehensive frames, but also those that are linked to the highest popularity. 

\section{Method}

In this section, we give an overview of the method proposed for video comixification. Our pipeline consists of two stages, described in the following sections. In Section 3.1 we describe in detail the keyframe extraction part, whereas in Section 3.2 we show how we implement the style transfer part. Fig.~\ref{pipeline-schema} shows a general architecture of the system.
As input our pipeline takes raw videos and some optional configuration parameters, and as output it produces full formated comics. In Section 4 we go into more detail about the real world implementation of the pipeline.

\begin{figure}[t!]
   \centering
\begin{tabular}{cc}

\includegraphics[width=7.5cm]{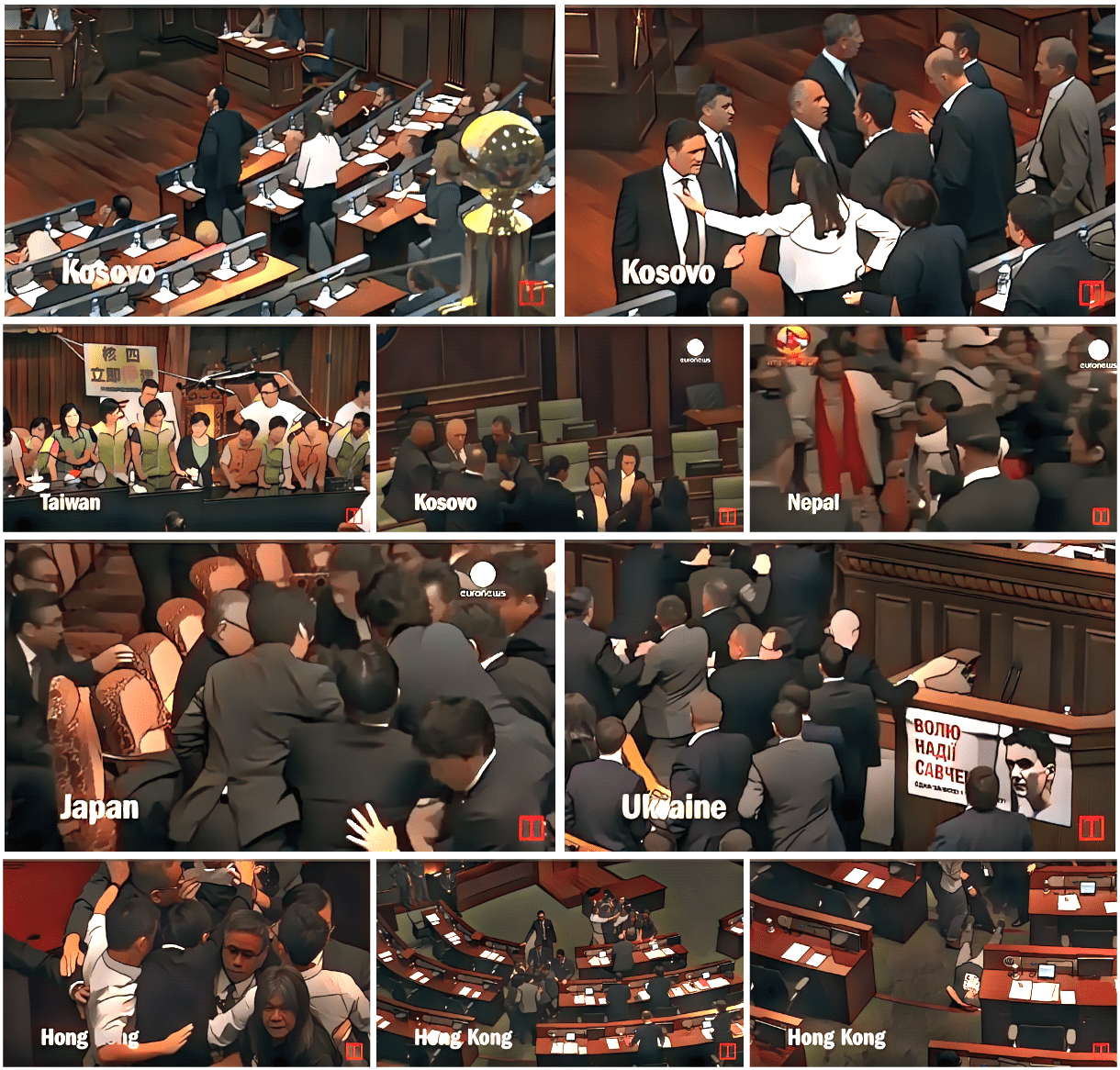}&
\includegraphics[width=7.5cm]{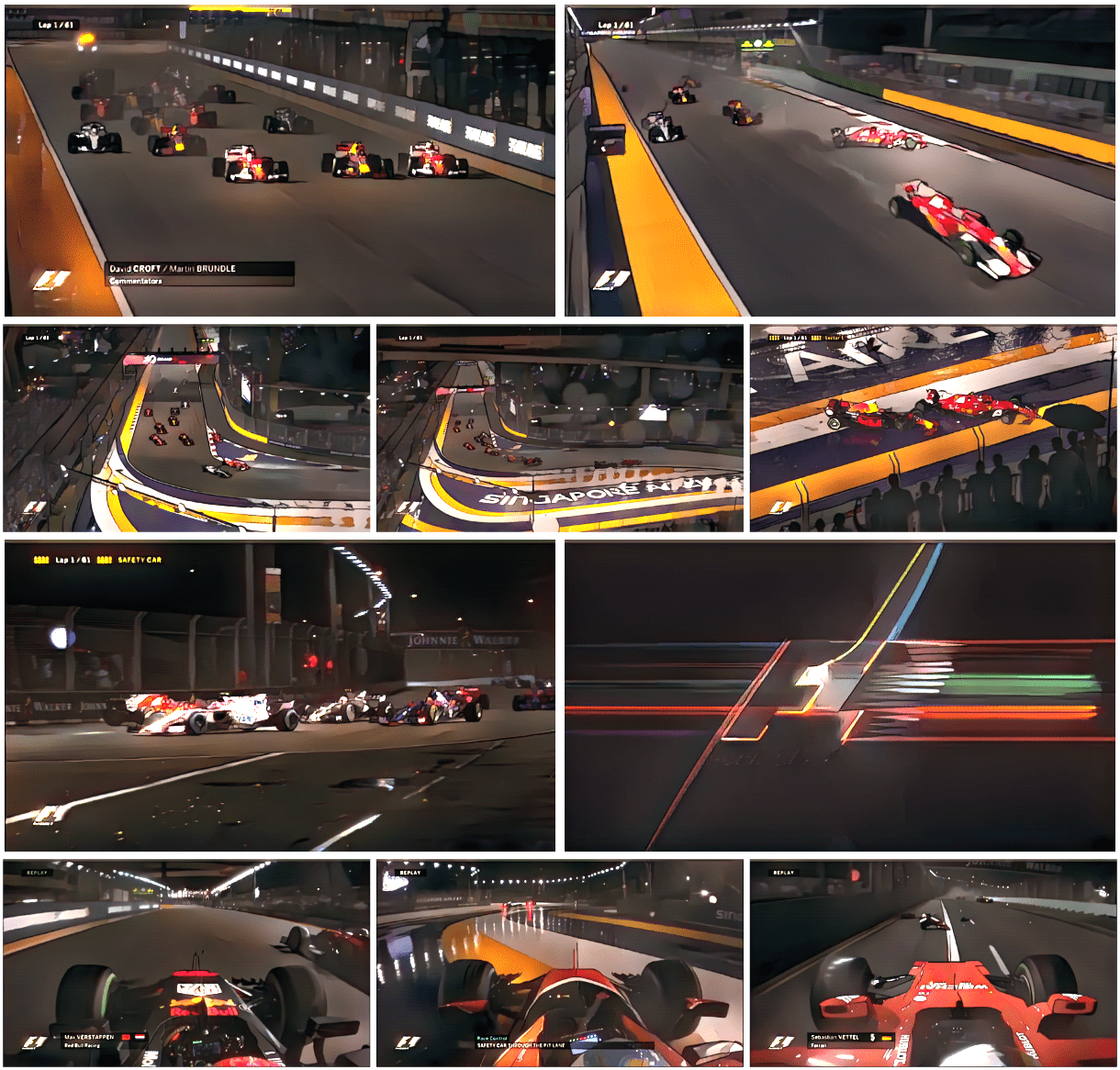}\\
a)&b)\\
\includegraphics[width=7.5cm]{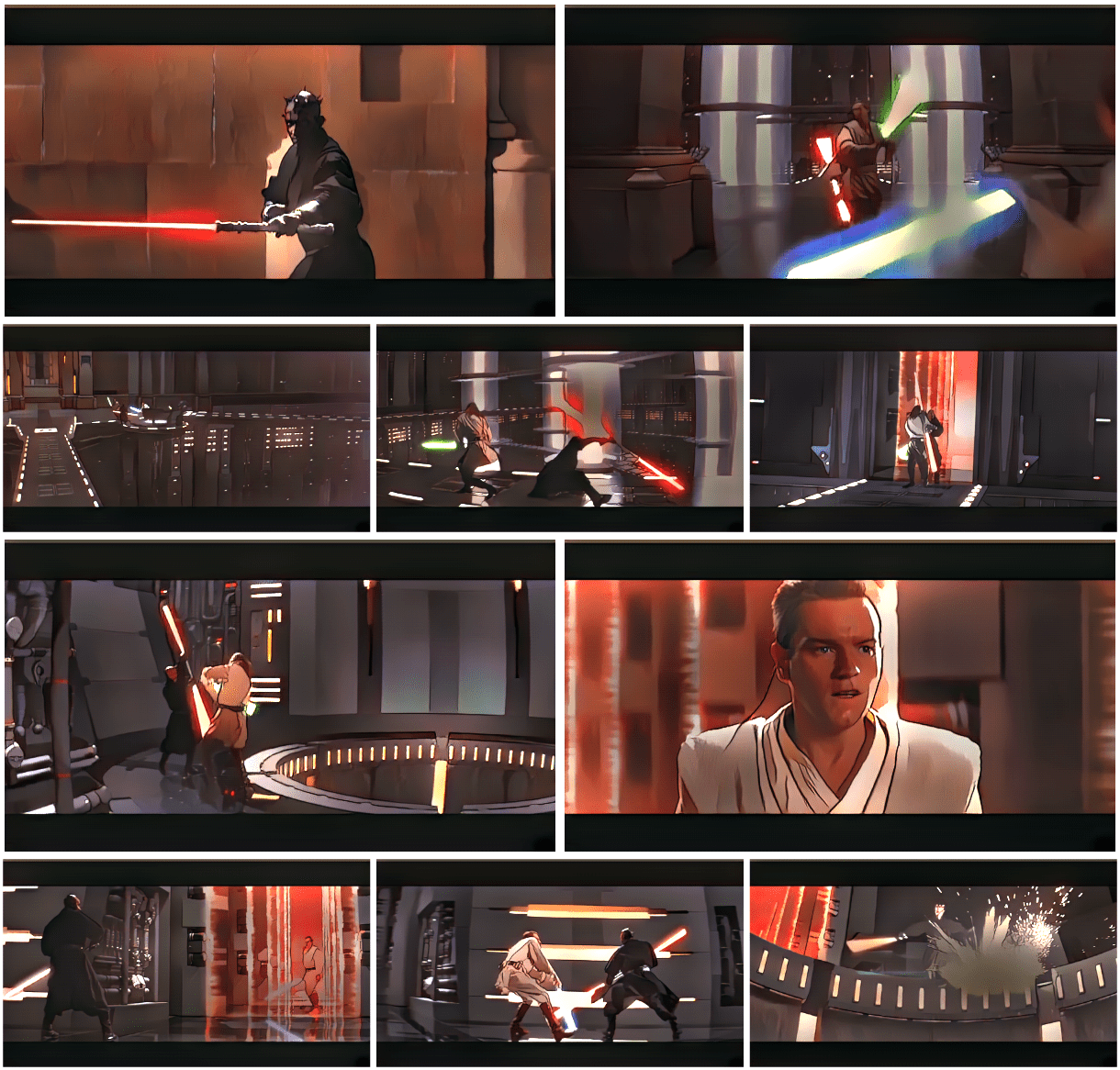}&
\includegraphics[width=7.5cm]{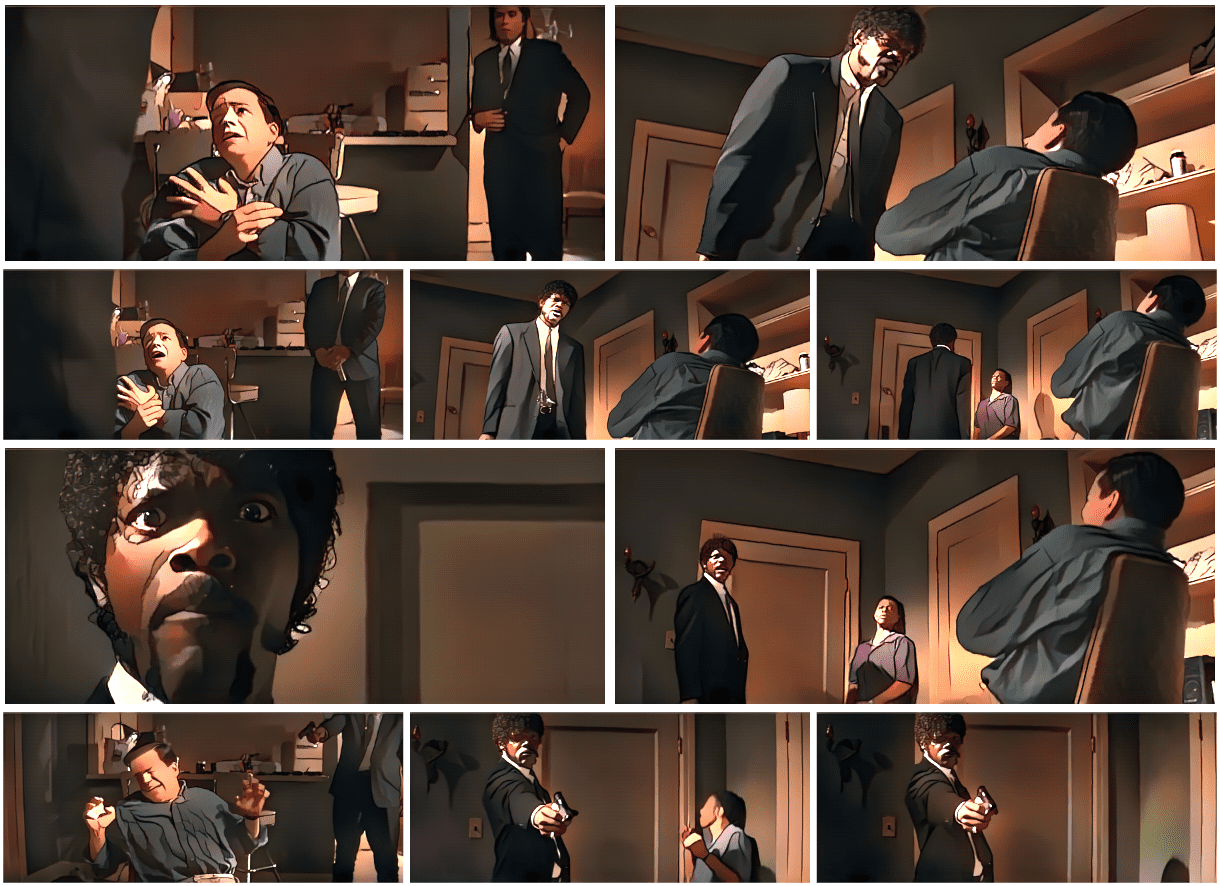}\\
c)&d)\\

\end{tabular}
    \caption{Examples of comixification results on different video types: a) political video, b) sport video, c) movie scene, d) movie scene. \label{comixify:results}}
\end{figure}

\subsection{Keyframes Extraction}

In this section, we describe the keyframes extraction process. The full process is accomplished by a combination of video summarization and image aesthetic estimation techniques. The task of video summarization is to create short and concise summaries of longer videos. The first step to create such summaries usually is to evaluate each frame’s highlightness, a score that will tell you how well the current frame represents the video. After having all frames evaluated for highlightness, most representative video segments are selected to be included in the summary. 

\begin{figure}[t!]
  \centering
  \includegraphics[width=0.9\textwidth]{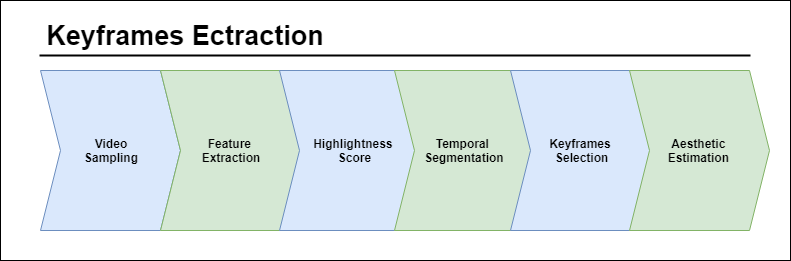}
  \caption{Keyframes extraction pipeline schema. The extraction consist of 6 stages: 1. Sampling of the video for frames to consider. 2. Features extraction from selected frames. 3. Computation of the highlightness score. 4. Temporal segmentation of video. 5. Keyframes selection using highlightness score and video segmentation. 6. Frames aesthetic estimation to reduce the number of selected keyframes.\label{keyframe-schema}}
\end{figure}

For our pipeline, we take the model from Zhou \textit{et al.} \cite{zhou2017reinforcevsumm} as a baseline and extend it with aesthetic estimation. Fig.~\ref{keyframe-schema} shows the full keyframe selection pipeline, divided into 6 steps.

For step 1 we follow the approach used in previous works \cite{zhou2017reinforcevsumm, zhang2016, mahasseni2017} on video summarization, where they sample videos with 2fps frequency.

In step 2 features are extracted from selected frames with GoogLeNet v1 \cite{googlenetv1} (pre-trained on ImageNet \cite{imagenet}) from the last pooling layer. The resulting feature vector is used in steps 3, 4, and 6.

In the following subsections we describe the other steps of our process.

\subsubsection{Highlightness score}

In step 3 we extract highlightness score. The model we use for that is an unsupervised RL model described by Zhou in \textit{et al.} \cite{zhou2017reinforcevsumm} and shown to outperform the competitors on the task of video summarization. In their work Zhou \textit{et al.} developed deep summarization network (DSN) that predicts for each frame a probability, which indicates how likely a frame is to be selected as a representative frame of the video. That probability is what we call the highlightness score. To train their DSN, they propose an end-to-end, reinforcement learning based framework. The reward function they use in their RL model is a function that jointly accounts for diversity and representativeness of DSN generated summaries. During training, the reward function judges how diverse and representative the generated summaries are, while DSN aims at earning higher rewards by learning to produce more diverse and more representative summaries. Since labels are not required, their method can be fully unsupervised. We train their model on the four datasets they used, and also additionally expand the training with the additional dataset, called VTW \cite{vtw}.

\subsubsection{Temporal segmentation}

In step 4 we split video into segments, and in step 5 we combine the forementioned segments with highlightness score to get a certain number of chosen frames. Segmentation is done mainly to further enforce the diversity of frames, as it limits the possibility of bordering frames being selected.

For temporal segmentation, we use KTS, proposed by Potapov \textit{et al.} \cite{potapov2014}, which segments a video into segments based on temporal differences in frames. We slightly modify the algorithm to be able to set a minimum constraint on number of segments $n$. We segment a video into $n' \geq n$. From which we select \textit{n} segments with highest highlightness by averaging the score from each frame in a segment. Then we find one frame in each of \textit{n} segments with the highest score. On these resulting \textit{n} frames we apply popularity estimation, to select the final \textit{k} keyframes.  

\subsubsection{Image aesthetic estimation}

Estimation of image aesthetic has been included in our pipeline as the last stage of keyframes selection - we limit the number of frames selected by temporal segmentation to the expected number \textit{k} (\textit{k $\leq$ n}, \textit{k}~is a divisor of \textit{n}) by arranging selected frames in order of their occurrence, dividing them into \textit{k} equal groups and selecting frame with the highest estimated aesthetic score in each group. 

\subsubsection{Image aesthetic - scoring}
Image aesthetic score can be obtained using one of two methods - either popularity estimation or image quality estimation:

\paragraph{Popularity estimation}
For popularity estimation we use baseline model from \cite{popularity}. We use the same dataset of over 37'000 thumbnails of videos published on Facebook. Every video has a calculated normalized popularity score, according to the following formula:
\begin{equation}
\textit{normalized popularity score} = log \frac{\textit{viewcount} + 1} {\textit{number of publisher's followers}}
\end{equation}

We train Support Vector Regression (SVR) model and evaluate it using Spearman correlation. We achieved the Spearman correlation coefficient value of 0.41. We use estimated popularity score as our aesthetic score.

\paragraph{Image quality estimation}
For image quality estimation we use approach proposed in NIMA: Neural Image Assesment \cite{talebi2018}. Model (based on NASNet-A \cite{zoph2017} architecture) is trained on AVA \cite{murray2012} dataset using squared EMD loss \cite{hou2016}. Output of this model is estimated histogram of ratings on scale from 1 to 10. We use mean of these scores as our aesthetic score.\

\subsection{Comics style transfer}

In this section, we first give an overview of the solutions proposed for transferring the style of images. We start with the initial work of Gatys {\it et al.}~\cite{gatys:} and continue with the most state-of-the-art methods with the best performance in terms of transfering comics style to images.

\subsubsection{Existing approaches}
Gatys \textit{et al.} \cite{gatys:} in their work, used 16 convolutional and 5 pooling layers of the 19-layer VGG network \cite{vgg:}. They passed three images through this network: Content image, Style image and White Noise image. The content information is given by feature maps from one layer and content loss is the squared-error loss between the two feature representations:
\begin{equation}
  L_{content} = \frac{1}{2} \sum_{i,j} (F_{ij}^{l} - P_{ij}^{l})^2 ,
\end{equation}

Where $F_{ij}^{l}$ is the activation of the i-th filter at position $j$ in layer $l$. On the other side, the style information is given by the Gram matrix $G$ of the vectorized feature maps $i$ and $j$ in layer $l$:
\begin{equation}
  G_{ij}^{l} = \sum_{k} F_{ik}^{l} F_{jk}^{l} ,
\end{equation}

The style loss is a squared-error loss function computed between two Gram matrices obtained from specific layers from white noise image and style image passed through the network. $N$ is a number of feature maps and $M$ is a feature map size.
\begin{equation}
  L_{style} = \frac{1}{2} \sum_{l=0}^L \omega_lE_l ,
\end{equation}
\begin{equation}
  E_{l} = \frac{1}{4N_l^2M_l^2} \sum_{i,j} (G_{ij}^{l} - A_{ij}^{l})^2 ,
\end{equation}

Finally, the cost function is defined as weighted sum of two above losses. Namely, between the activations (content) and Gram matrices (style) and then is minimized using backpropagation.
\begin{equation}
  L_{total} = \alpha*L_{content} + \beta*L_{style}
\end{equation}

Unfortunately, this method is very slow. Using this approach, style transfer of one 512x512 pixel image lasts almost a minute on recent GPU architectures, such as NVIDIA Quadro M6000 or Titan X~\cite{johnson:,demystif:}. It is caused by the fact that for each pair of images, this method performs an optimization process using backpropagation to minimize $L_{total}$. To address this shortcoming, several approaches have been proposed. M. Pęśko and T. Trzciński in \cite{neuralstyle-case-study} presented very detailed comparison of all state-of-the-art neural style transfer approaches in terms of comics style transfer. Two of them were found the best, namely Adaptive Instance Normalization approach presented by X. Huang and S. Belongie in their paper \cite{huang:belongie} and Universal Style Transfer proposed by Y. Li \textit{et al.} in \cite{universal:style}.

\subsubsection{Adaptive Instance Normalization}
This is the first fast and arbitrary neural style transfer algorithm that resolves problem with generalization, quality and efficiency trade-off. This method consists of two networks: a style transfer network and a loss network. 

\begin{figure}[t!]
  \centering
  \includegraphics[width=0.9\textwidth]{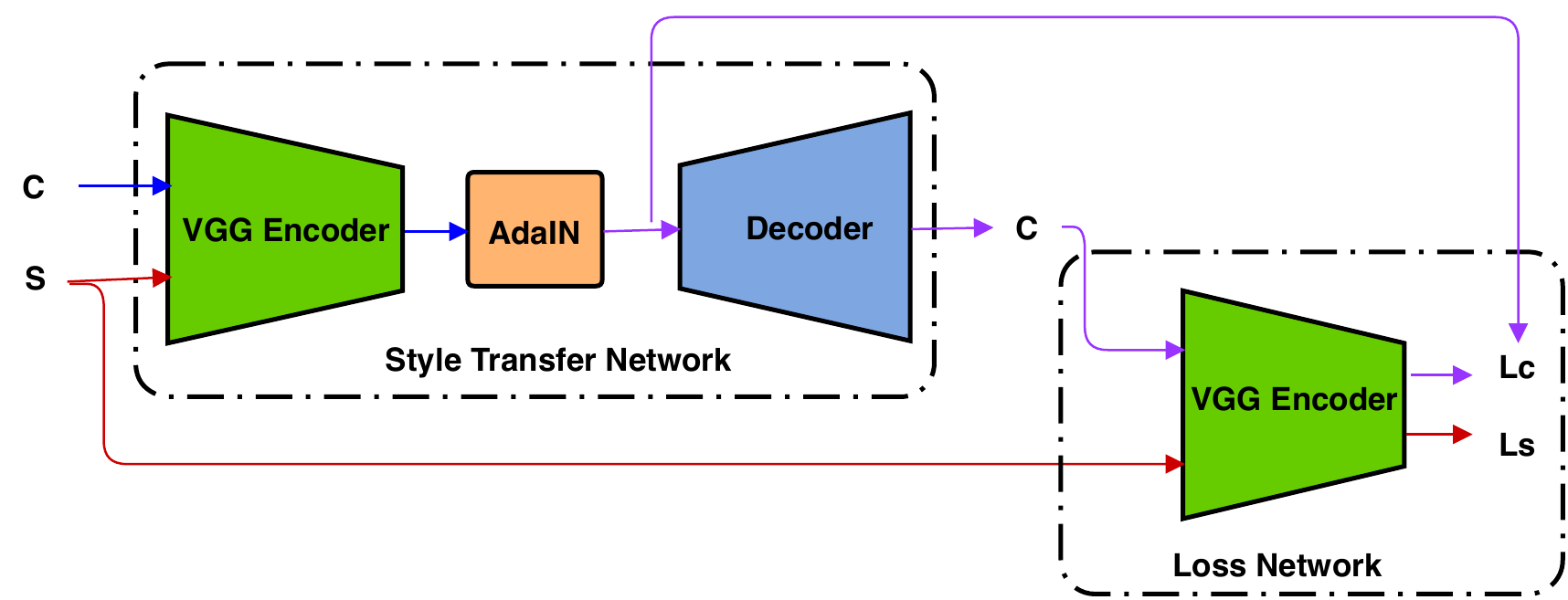}
  \caption{AdaIN architecture. C and S are Content and Style images and Lc and Ls are content and style losses. \label{AdaIN:arch}}
\end{figure}

The loss network is pretty similar to the network presented in~\cite{gatys:}. It is used to compute a total loss which is minimized by a backpropagation algorithm to fit the parameters of a style transfer network. The style transfer network consists of a simple encoder-decoder architecture. The encoder is first few layers of a pre-trained VGG-19 network. The decoder mirrors the encoder with a few minor differences. All pooling layers are replaced by nearest up-sampling layers and there are no normalization layers. The most interesting part is AdaIN layer which is placed between the encoder and the decoder. AdaIN produces the target feature maps that are inputs to the decoder by aligning the mean and variance of the content feature maps to style feature maps. A randomly initialized decoder is trained using a loss network to map outputs of AdaIN to the image space in order to get a stylized image. AdaIN architecture can be seen in the Fig. \ref{AdaIN:arch}. 

\subsubsection{Universal Style Transfer}

Knowing that covariance matrix can be as effective as Gram matrix in representing image style \cite{cov:gram}, Y. Li \textit{et al.} \cite{universal:style} proposed a novel approach called Universal Style Transfer (UST). It is closely related to AdaIN method but the intuition of UST is to match covariance matrices of feature maps instead of aligning only mean and variance which was proposed in \cite{huang:belongie}. Their Universal Style Transfer approach (UST) uses a very similar encoder-decoder architecture where encoder is composed of a first few layers of a pre-trained VGG-19 network and the decoder mostly mirrors the encoder. However, instead of using the AdaIN layer to carry style information, they used Whitening and Coloring transform (WCT). Moreover, in this method style transfer that is represented by WCT layer is not used during training. The network is only trained to reconstruct an input content image. The entire style transfer takes place in the WCT layer which is added to already pre-trained image reconstruction network. 

\begin{figure}[t!]
  \centering
  \includegraphics[width=0.7\textwidth]{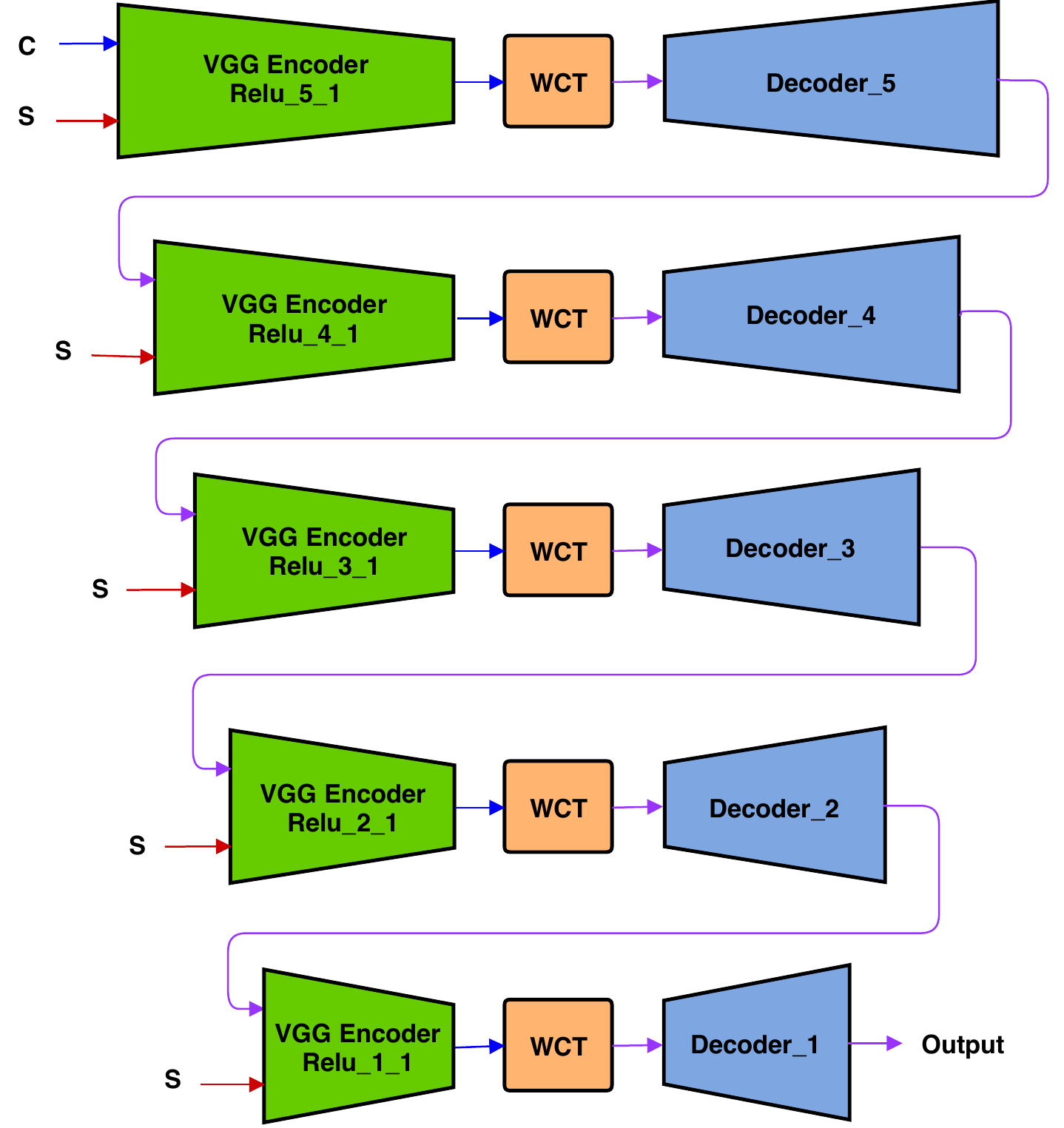}
  \caption{Universal Style Transfer architecture with the whole multi level pipeline. Each level of stylization consists of single encoder-WCT-decoder network with different decreasing number of VGG layers. C and S are content and style images, respectively. \label{ust:arch2}}
\end{figure}

Another extension is to use a multi-level stylization in order to match the statistics of the style at all abstraction levels. It means that the result obtained from a network that matches higher level information is treated as the new content input image to network that matches lower level statistics. Such architecture can be seen in Fig. \ref{ust:arch2}. 

\begin{figure}[t!]
\centering
\bgroup
\setlength\tabcolsep{0.2em}
\begin{tabular}{cccc}

\includegraphics[width=3.8cm]{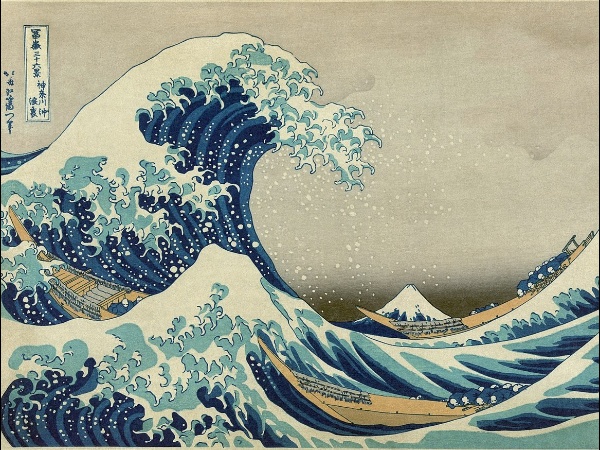}&
\includegraphics[width=3.8cm]{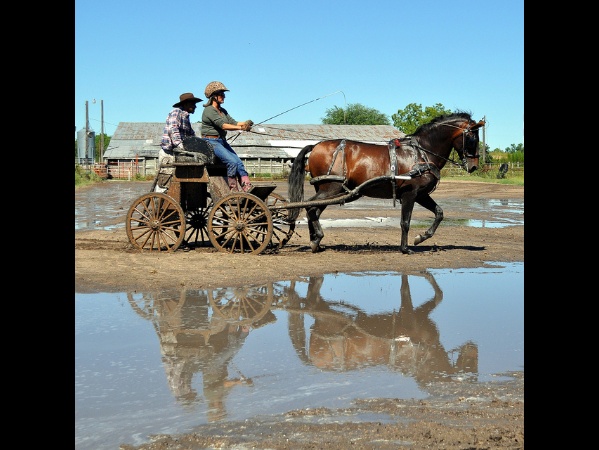}&
\includegraphics[width=3.8cm]{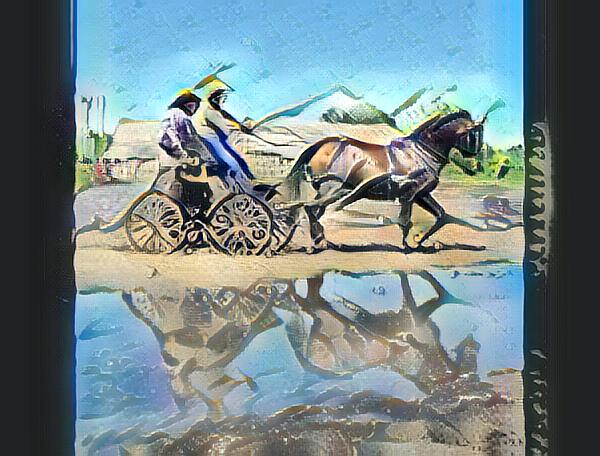}&
\includegraphics[width=3.8cm]{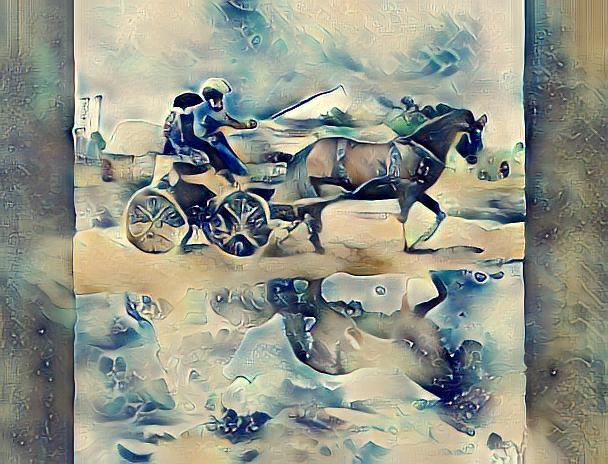}\\

\includegraphics[width=3.8cm]{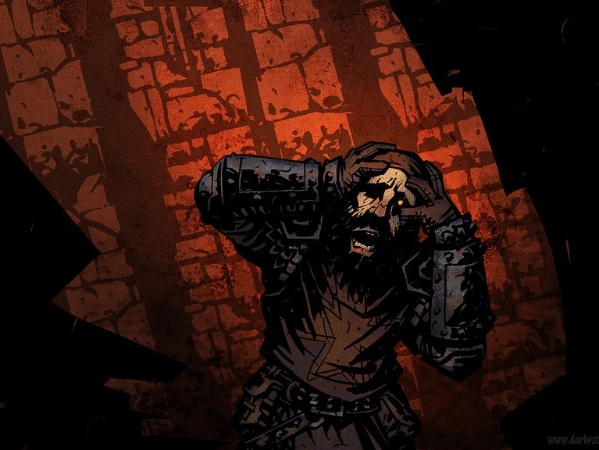}&
\includegraphics[width=3.8cm]{Content/3.jpg}&
\includegraphics[width=3.8cm]{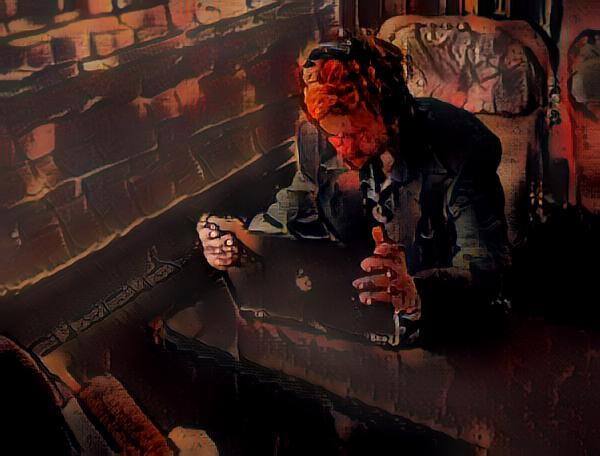}&
\includegraphics[width=3.8cm]{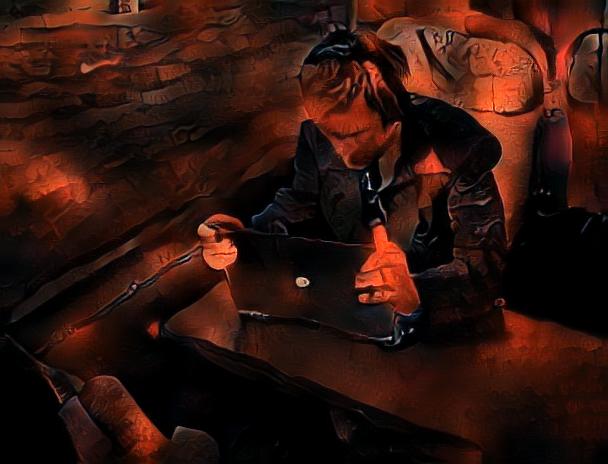}\\

\includegraphics[width=3.8cm]{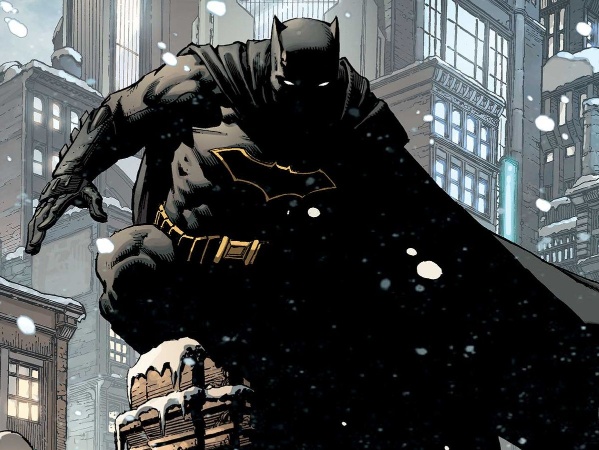}&
\includegraphics[width=3.8cm]{Content/6.jpg}&
\includegraphics[width=3.8cm]{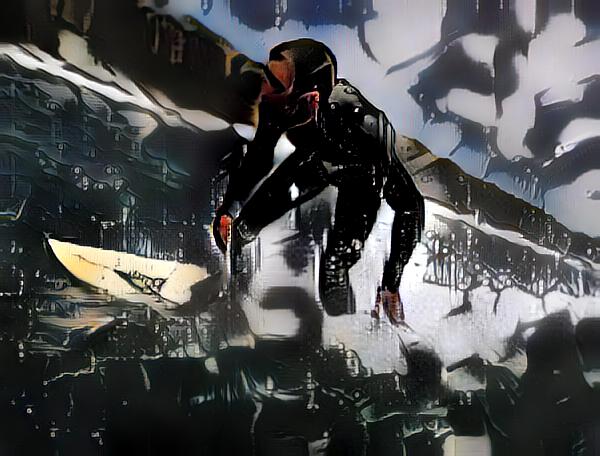}&
\includegraphics[width=3.8cm]{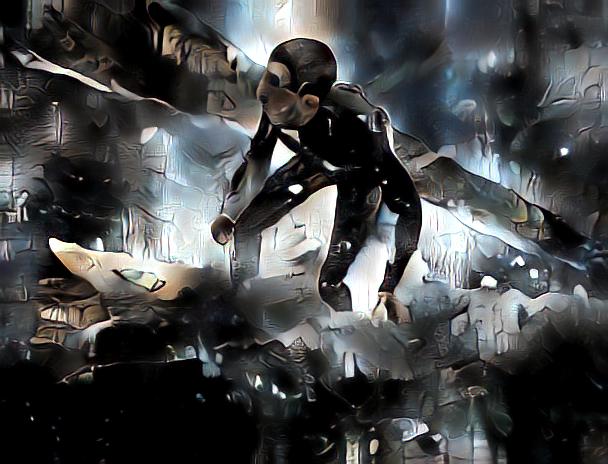}\\

\includegraphics[width=3.8cm]{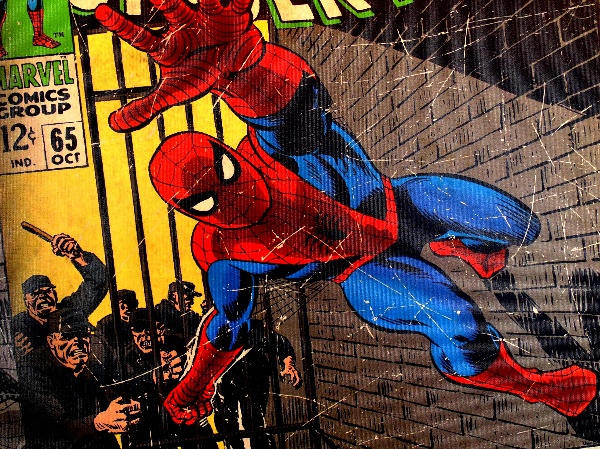}&
\includegraphics[width=3.8cm]{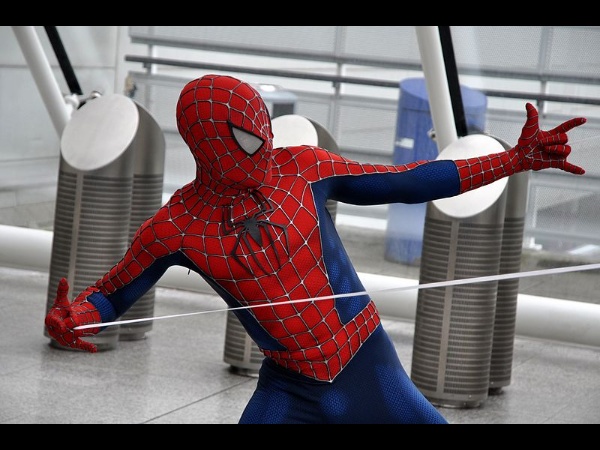}&
\includegraphics[width=3.8cm]{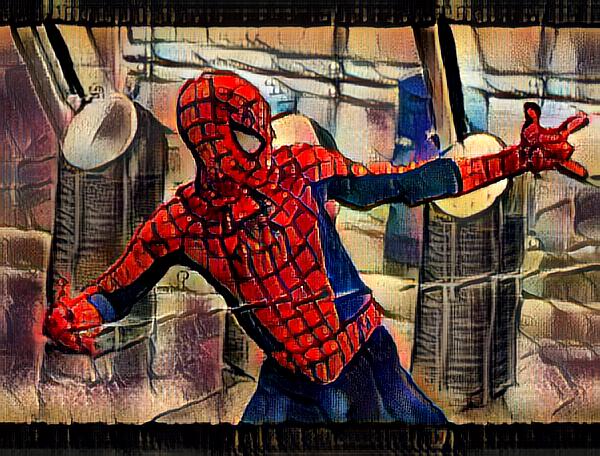}&
\includegraphics[width=3.8cm]{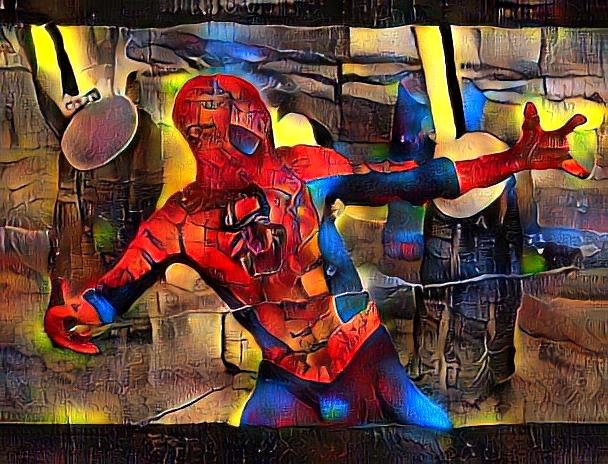}\\

a) Style images &b) Content images & c)AdaIN &d) UST-WCT\\
\end{tabular}
    \caption{Results from different style transfer methods. \label{visual:results}}
\egroup
\end{figure}

Sample results presenting above-mentioned architectures can be found in Fig.~\ref{visual:results}. Unfortunately, each of those two methods suffers from some common problems such as inappropriate color transfer and blur effects. AdaIN method returns images that are close to cartoons, but it often leaves the content colors or some mixed hues that do not fit the style. UST-WCT give results with more appropriate and stylistically coherent colors. However, the results of those models seem to be stylized too much which often leads to significant distortions in those pictures. Moreover, for our "comixification" purpose we would like to have some more powerful solution that would be able to "remember" how to  apply comics style to any content image without need to give it any style image. Fortunately, recently, Chen, Yang et al. in \cite{cartoongan} presented completely different approach based on Generative Adversarial Networks that gives really promising results.

\subsubsection{CartoonGAN}
\begin{figure}[t!]
  \centering
  \includegraphics[width=\textwidth]{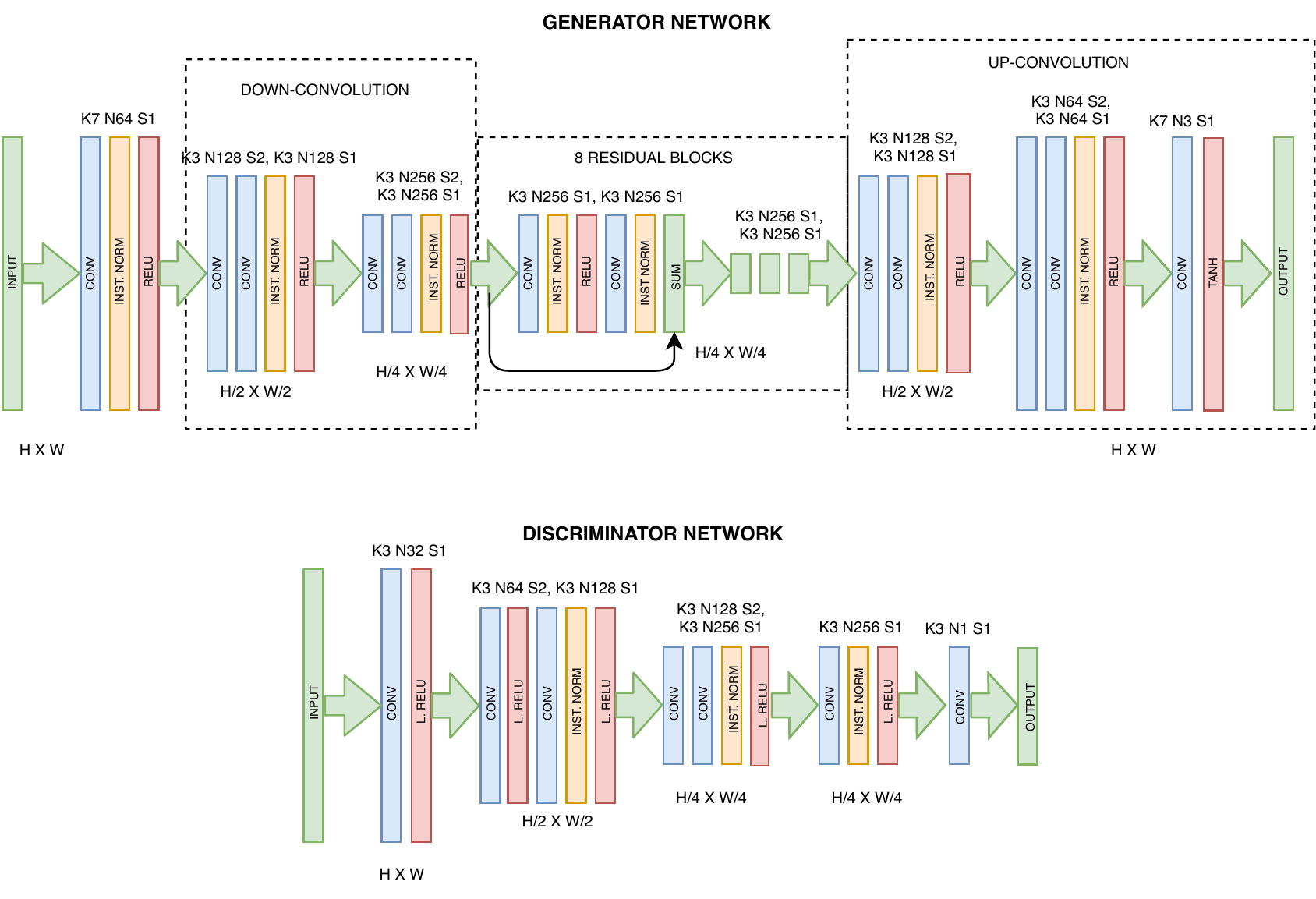}
  \caption{CartoonGAN architecture. K is the kernel size, N is the number of feature maps, S is the stride in each convolutional layer, 'INST. NORM' is an instance normalization layer, 'SUM' is an elementwise sum and 'L. RELU' is a leaky ReLu layer .\label{cartoongan-arch}}
\end{figure}

This approach faces Style Transfer problem in completely different way using so-called Generative Adversarial Networks (GANs). This GAN based method consists of two Convolution Neural Networks: Discriminator D and Generator G. Generator is trained to create images that fools the discriminator. On the other hand, the discriminator is trained to classifies whether the image is from the real target domain or synthetic, produced by Generator. G and D architecture can be seen in Fig. \ref{cartoongan-arch}. The Discriminator is designed to be shallow, because judging  whether image is a cartoon or not is less demanding task and it should rely only on local features of the image.

\bigskip
\textbf{Loss function}
The loss function $L(G,D)$ consists of 2 parts: the adversarial loss $ L_{adv}(G,D) $ that is responsible for achieving the desired style transformation by the generator network and the content loss $ L_{con}(G,D) $ which helps to persist the image content during training.  $ \omega $ is a parameter to balance between these two losses.

\begin{equation}
  L(G,D) = L_{adv}(G,D) + \omega*L_{con}(G,D)
\end{equation}

The adversarial loss is applied to both networks G and D. In comparison to classic GAN frameworks adversarial loss, authors presented so-called Edge Promoting Loss. What it means is to introduce additional class that represents cartoons with blurred edges and treat it like non-cartoon images. This extra addition handles the problem of clear edges that are very characteristic for cartoon/comics images and may not be recognize by Generator because of the small proportion of these edges in the whole image. The content loss is just L1 norm between features maps obtained from one layer from VGG network \cite{vgg:}.

\bigskip
\textbf{Initialization phase}
To improve GAN framework convergence, authors propose additional step before the actual training. This Initialization phase is a short training of the Generator network to reconstruct the input image. Training is performed using only content loss $L_{con}(G)$.

\subsubsection{ComixGAN}
However, CartoonGAN results are still not perfect. Authors shared two pretrained models, that are trained to output image in a style of two Anime artists Mamoru Hosoda and Miyazaki Hayao. This fact causes that output images are targeted to Anime style, which can be seen to have unnatural colors. Moreover, edges are still not so sharp and distinct, as they should be. Bearing in mind that we would like to get very universal comic images with more natural and contrasting colors and more distinct edges we decide to train our own ComixGAN framework based on CartoonGAN approach.

\bigskip
\textbf{Data}
For training we use real, comics and comics with blurred edges images as described in \cite{cartoongan}. As real image we use MS COCO dataset \cite{ms-coco}. For comics images we use keyframes obtained from different cartoons that can be described as being comic book style.

\bigskip
\textbf{Architecture}
We start with identical architecture and parameters to original CartoonGAN framework, but surprisingly results turn out to be poor. Generated images lack the comic book style and have no distinct edges. Moreover, networks are very unstable during the training. To mitigate those shortcomings of the base model, we introduce the following changes:
\begin{itemize}
\item As a Generator loss function we use non-saturating loss (as described in \cite{ian:gan})  instead of standard minimax loss version.
\item We use sigmoid function in the last layer of Discriminator Network.
\item We use Generator/Discriminator ratio in training equals to 3:1 (Three updates of Generator weights per one update of Discriminator)
\item We use initial training also for Discriminator to pretrain it to distinguish comics images from real and edge-blurred comics.
\end{itemize}

\begin{figure}[t!]
\centering
\bgroup
\setlength\tabcolsep{0.2em}
\begin{tabular}{cccc}

\includegraphics[width=3.8cm]{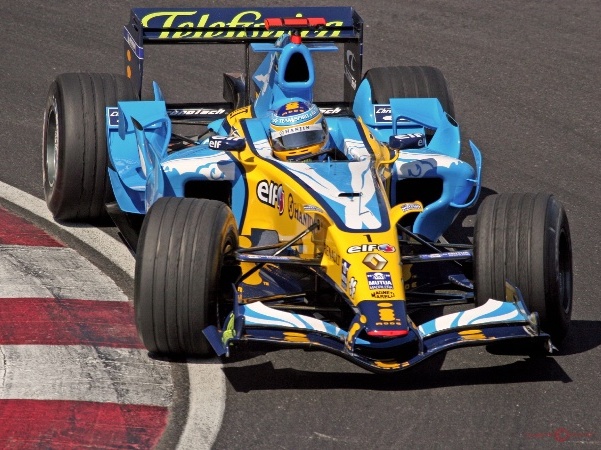}&
\includegraphics[width=3.8cm]{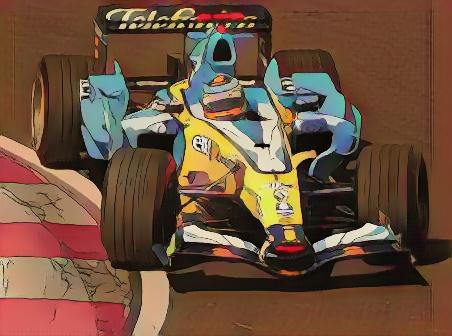}&
\includegraphics[width=3.8cm]{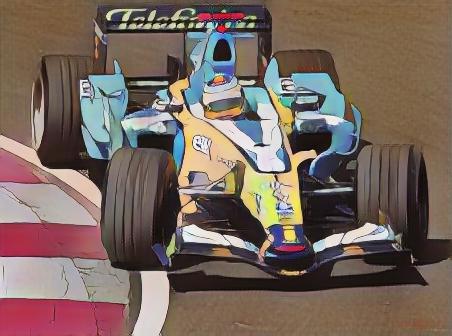}&
\includegraphics[width=3.8cm]{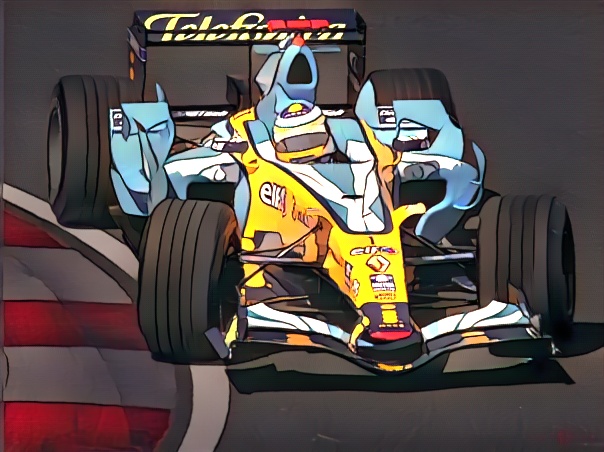}\\

\includegraphics[width=3.8cm]{Content/3.jpg}&
\includegraphics[width=3.8cm]{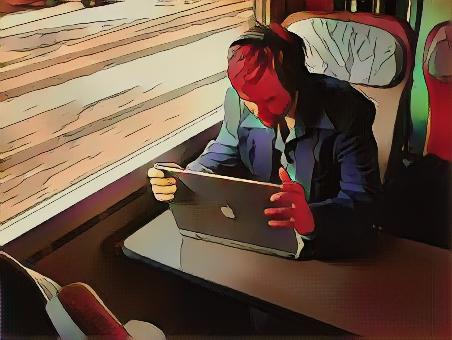}&
\includegraphics[width=3.8cm]{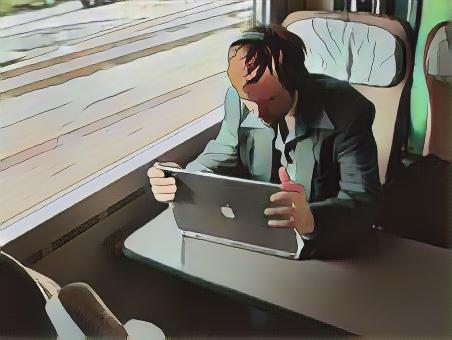}&
\includegraphics[width=3.8cm]{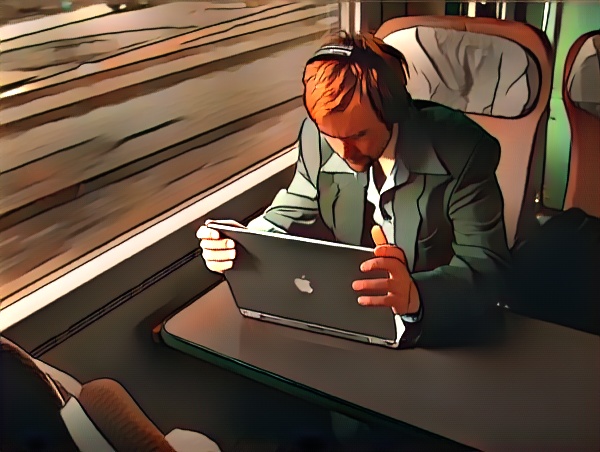}\\

\includegraphics[width=3.8cm]{Content/7.jpg}&
\includegraphics[width=3.8cm]{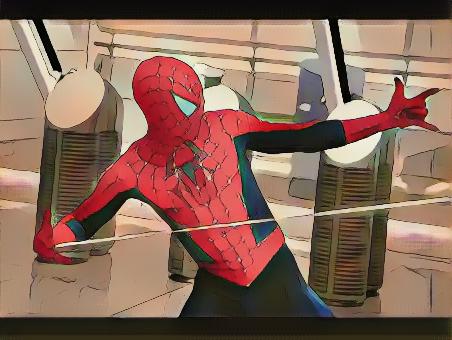}&
\includegraphics[width=3.8cm]{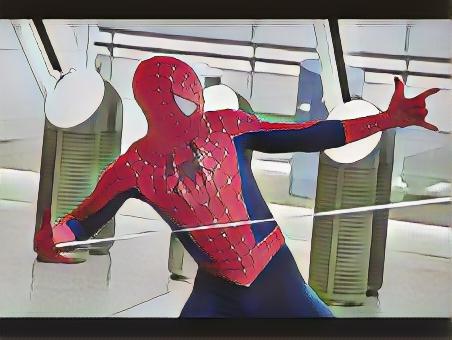}&
\includegraphics[width=3.8cm]{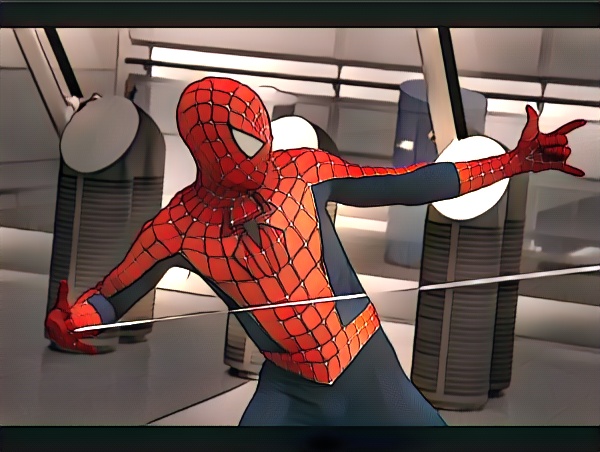}\\

\includegraphics[width=3.8cm]{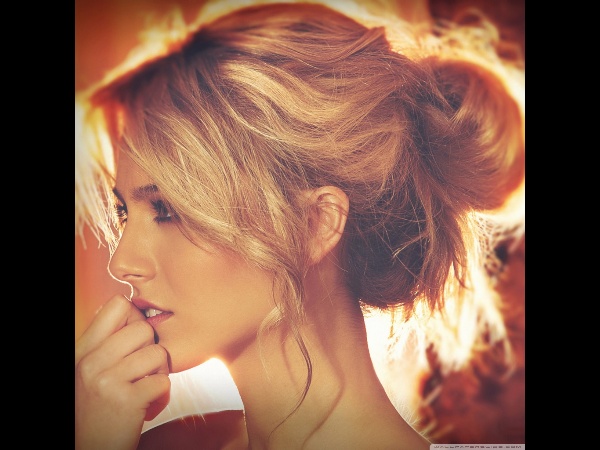}&
\includegraphics[width=3.8cm]{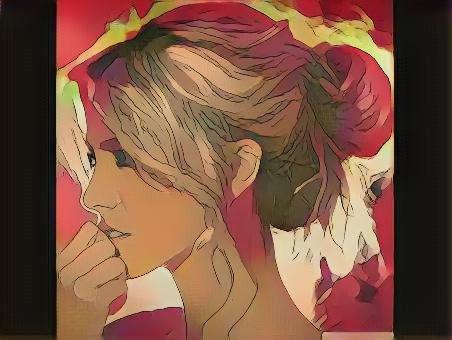}&
\includegraphics[width=3.8cm]{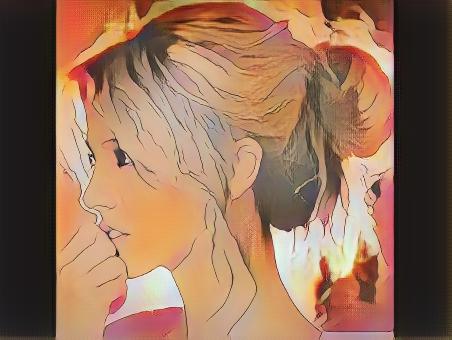}&
\includegraphics[width=3.8cm]{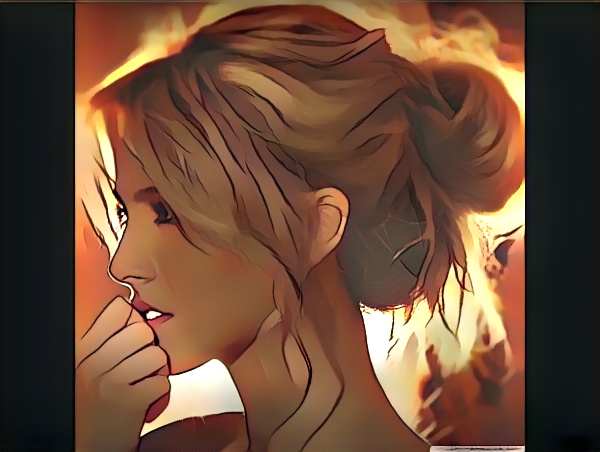}\\

a)&b)&c)&d)\\
\end{tabular}
    \caption{Results for different comixification approaches. Columns: a) content images, b) original CartoonGAN model (Hayao), c) original CartoonGAN model (Hosoda), d) ComixGAN.
    \label{comixgan-results}}
\egroup
\end{figure}

In Fig.~\ref{comixgan-results} we present some sample results obtained from original CartoonGAN framework in both Hayao and Hosoda styles (columns b) and c)) and our ComixGAN framework with improvements that we introduce to obtain better quality results (column d)). We can see that our ComixGAN produces images with more distinct and clear edges than all remaining approaches. Moreover, colors are uniform and vivid which is very characteristic for comics style. Also content information in preserved very good with no visible distortions. The quality of images produced by ComixGAN seems to be better than in all previous approaches. This fact prevailed and we decide to use is as a main style transfer method in our comixify pipeline. On the other hand, to give everyone the opportunity to compare, in our application we also include the option of choosing between the models provided by CartoonGAN authors.

\section{Web application}

\begin{figure}[t!]
  \centering
  \includegraphics[width=0.7\textwidth]{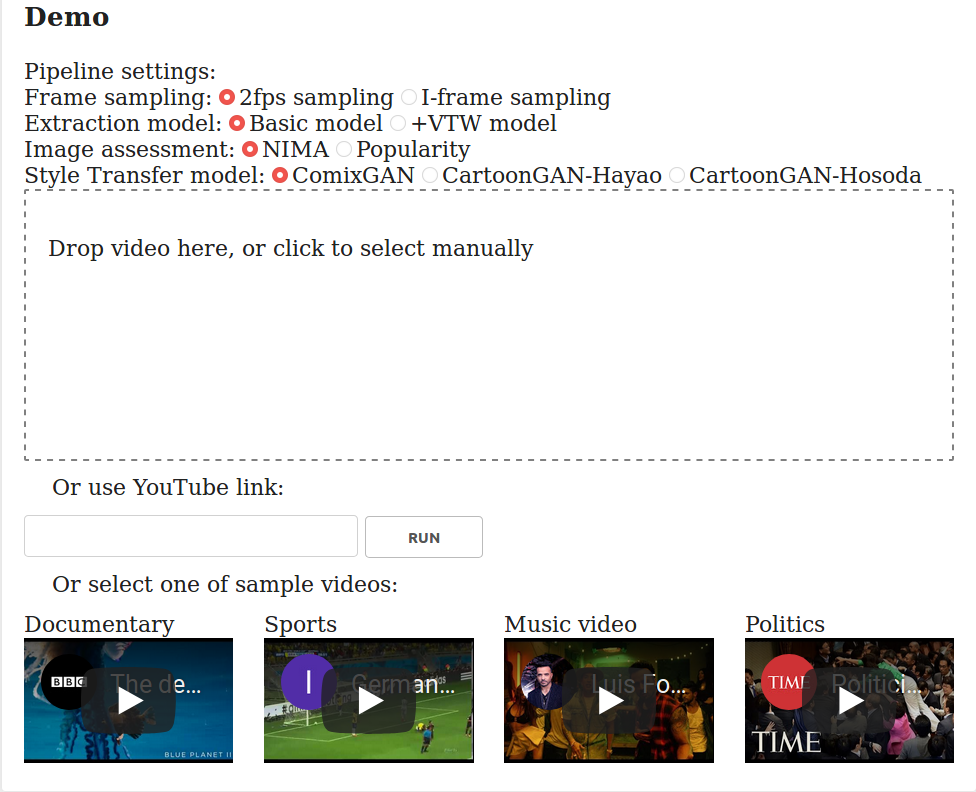}
  \caption{Comixify web application screen-shot. \label{app-screenshot}}
\end{figure}

In order to demonstrate our solution, we share a Web-based application that allows us to process video files using our pipeline. Our application is publicly available at \url{https://comixify.ii.pw.edu.pl/}. Frontend interface can be also seen in Fig. \ref{app-screenshot}. We publish the source code of implementation at \url{https://github.com/maciej3031/comixify}. The application gives everyone the opportunity to test our solution in three ways: by uploading some video file, by providing a link to the YouTube video or by selecting a video from available samples. One can choose which image aesthetic estimation model will be used - popularity estimation or image quality estimation (NIMA). The application also allows to select frame extraction model to experiment with results. There is a choice between Basic model and Basic + VTW model. What is more, also Style Transfer model can be chosen. One can select between ComixGAN model and both CartoonGAN models \cite{cartoongan}. In addition, a REST API is also available, allowing the use of the pipeline without necessity of frontend experience.

\section{Conclusions}

In this paper, we introduced an extensive pipeline for transforming videos into comics. Our solution works in semi real time manner and produces convincing, eye-pleasing comics layouts. Our approach can be extended by adding some generative comics layout composition and by introducing voice recognition to add text annotation to the comics. This is a direction we plan to explore in the future work.

\bibliography{citations3}

\end{document}